\documentclass[12pt]{article}

\usepackage{xcolor}
\newcommand{\wallpapergroup}[2]{{$\textcolor{blue}{#1}\textcolor{red}{#2}$}}

\usepackage{scicite}

\usepackage{siunitx}
\usepackage{url}
\usepackage{times}
\usepackage{amsfonts}
\usepackage{siunitx}
\usepackage{algorithm}
\usepackage{algpseudocode}
\usepackage{graphicx}
\usepackage{amsmath}
\usepackage[labelformat=empty]{caption}
\usepackage{gensymb}

\newcounter{myparagraph}

\newenvironment{sciabstract}{%
\begin{quote} \bf}
{\end{quote}}

\title{Bridging Hard and Soft: Mechanical Metamaterials Enable Rigid Torque Transmission in Soft Robots}

\author
{Molly Carton,$^{1,2 \dagger}$ Jakub F. Kowalewski,$^{3 \dagger}$ Jiani Guo$^{4}$, Jacob F. Alpert$^{3}$, Aman Garg,$^{4}$ \\ Daniel Revier,$^{5}$ and Jeffrey Ian Lipton$^{3\ast}$ \\
\\
\normalsize{$^{1}$Department of Mechanical Engineering, Massachusetts Institute of Technology}\\
\normalsize{77 Massachusetts Ave., Cambridge, MA 02139, USA}\\
\normalsize{$^{2}$Department of Mechanical Engineering, University of Maryland}\\
\normalsize{4289 Campus Dr., College Park, MD 20742, USA}\\
\normalsize{$^{3}$Department of Mechanical and Industrial Engineering, Northeastern University}\\
\normalsize{390 Huntington Ave., Boston, MA 02115, USA}\\
\normalsize{$^{4}$Department of Mechanical Engineering, University of Washington}\\
\normalsize{3900 E Stevens Way NE, Seattle, WA 98195, USA}\\
\normalsize{$^{5}$Department of Computer Science, University of Washington}\\
\normalsize{185 E Stevens Way NE, Seattle, WA 98195, USA}\\
\normalsize{$^\ast$To whom correspondence should be addressed; E-mail:  j.lipton@northeastern.edu} \\
\normalsize{$^\dagger$These authors contributed equally to this work.}
}

\date{}

\begin{document} 


\baselineskip24pt


\maketitle


\begin{sciabstract}
Torque and continuous rotation are fundamental methods of actuation and manipulation in rigid robots. 
Soft robot arms use soft materials and structures to mimic the passive compliance of biological arms that bend and extend.
This use of compliance prevents soft arms from continuously transmitting and exerting torques to interact with their environment.
Here, we show how relying on patterning structures instead of inherent material properties allows soft robotic arms to remain compliant while continuously transmitting torque to their environment. We demonstrate a soft robotic arm made from a pair of mechanical metamaterials that act as compliant constant-velocity joints. 
The joints are up to 52 times stiffer in torsion than bending and can bend up to 45$\degree$. This robot arm can continuously transmit torque while deforming in all other directions. The arm's mechanical design achieves high motion repeatability (0.4 mm and 0.1$\degree$) when tracking trajectories. We then trained a neural network to learn the inverse kinematics, enabling us to program the arm to complete tasks that are challenging for existing soft robots such as installing light bulbs, fastening bolts, and turning valves. The arm's passive compliance makes it safe around humans and provides a source of mechanical intelligence, enabling it to adapt to misalignment when manipulating objects.
This work will bridge the gap between hard and soft robotics with applications in human assistance, warehouse automation, and extreme environments.

\end{sciabstract}

\noindent \textbf{One Sentence Summary} We present a soft robot arm that can exert continuous torques based on a mechanical metamaterial.


\section*{INTRODUCTION}

\begin{figure}[!h]
\makebox[\textwidth][c]{\includegraphics[width=\textwidth]{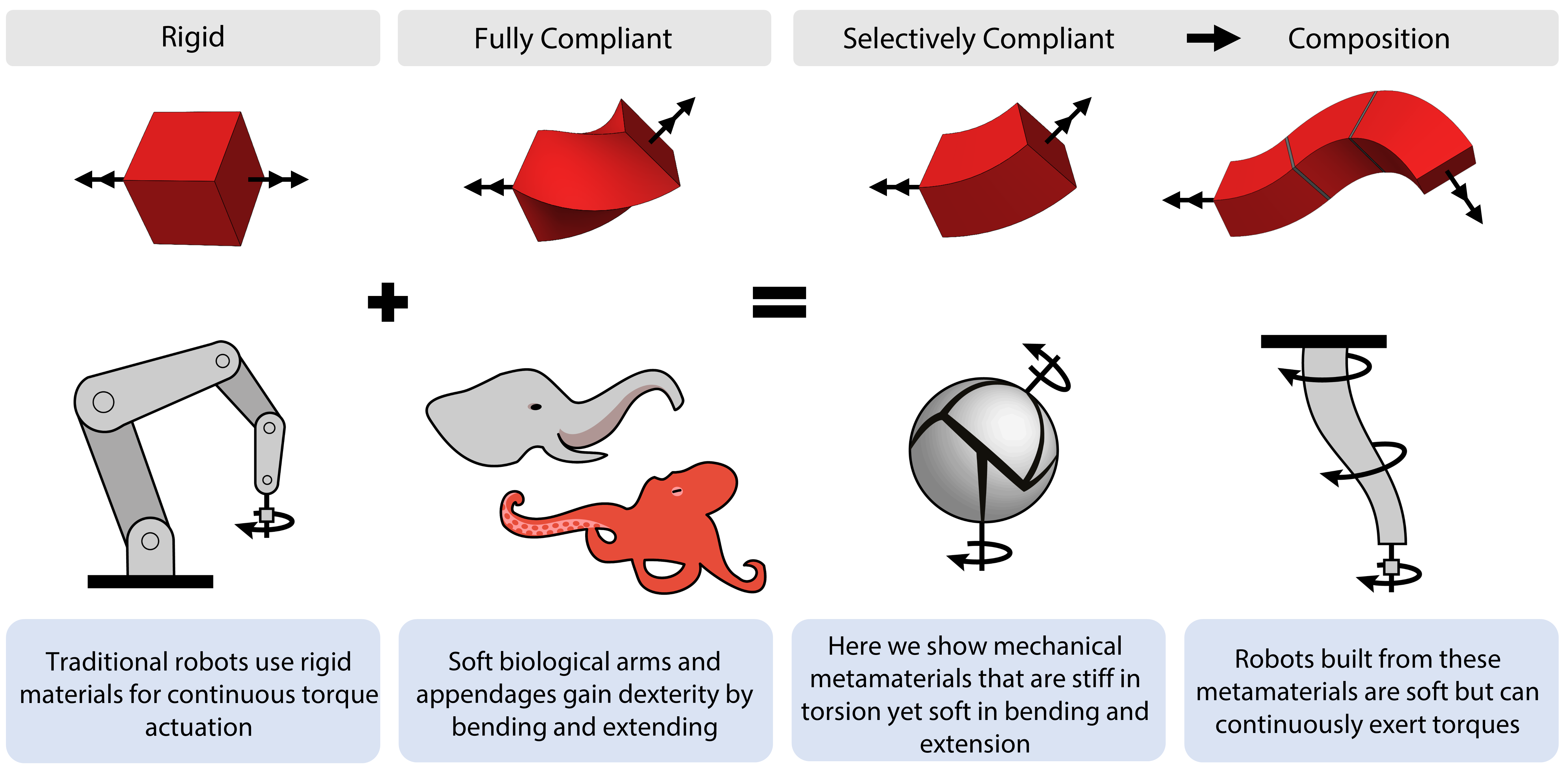}}
\caption{{Fig. 1. \bf TRUNCs conceptual overview.} We present a joint that is torsionally rigid yet compliant in bending and extension. The joint can be directly connected to a motor to create soft torque actuators. Multiple joints can then be composed to create soft arms with continuous torque actuation.}
\label{fig:Concept}
\end{figure}

Soft robots leverage their distributed compliance to interact safely with humans \cite{Guan2023-gj} and the environment \cite{Del_Dottore2024-qg,Hawkes2017-tl}. However, entirely compliant robots struggle to twist objects and transmit continuous rotations. In applications such as reorienting parts in warehouse automation \cite{correll2016analysis}, fastening bolts during industrial assembly \cite{Hamdule2024-ai}, or twisting multi-turn valves in extreme human-engineered environments \cite{Krotkov2018-nv}, soft robots remain unequipped to exert the necessary torque to manipulate their environment. For traditional robot arms, torque can be directly and continuously transmitted through links due to their torsional stiffness. While this makes rigid robots efficient at twisting tasks, they carry inherent safety risks due to high mass and stiffness, resulting in robots working in structured environments isolated from humans. By mimicking soft biological appendages such as elephant trunks \cite{Hannan2003-mj,grzesiak2011bionic,Guan2023-gj,Guan2020-nl,Zhang2023-kz} and octopus tentacles \cite{Immega1995-kq,McMahan2006-ks,Cianchetti2012-ru,Pettinato1989-pn,Han2022-ku}, soft robot arms are not only safe around humans but can also move with high dexterity and adapt to environmental uncertainty. Bio-inspired robots that combine bending with extension gain additional agility over just bending \cite{Schulz2022-dd,shikari2018triple}. We need robot arms that combine the safety and dexterity of bending and extending with the utility of continuous torque transmission to work directly alongside humans. 

Building soft robots from soft materials makes torque transmission inherently difficult. Materials such as elastomers are commonly used to build soft arms \cite{Calisti2011-vc,Martinez2013-wd}, but lack the torsional rigidity necessary for exerting torques.  Existing flexible torque couplings, such as flex shafts and bellows, have significant limitations. Flex shafts have a high torsional rigidity and can be directly embedded within soft robots \cite{Tan2018-dj}, however, they are only compliant in bending, and thus restrict the robot from extending. Bellows, while able to bend and extend \cite{Kim2024-nd}, are difficult to nest and must sacrifice their range of motion for torsional stiffness. Alternatively, twist actuation can be achieved by introducing chirality into a soft structure. Soft robots that use pneumatic artificial muscles (PAMs) with angled fibers \cite{Connolly2017-ph,Bishop-Moser2015-np}, inflatable helical structures \cite{Yan2018-ht,Blumenschein2018-it,Oh2023-xe}, vacuum powered elastomeric twist actuators \cite{Yang2015-qi,Jiao2019-ft,Chen2021-qd}, foldable origami \cite{Liu2018-iy,Li2023-gw}, and handed shearing auxetics (HSAs) \cite{lipton632,chin2018compliant} can transmit rotation but are all confined to a finite range of motion. These existing solutions cannot be continuously driven to transmit multiple torques while remaining flexible in extension and bending.

Creating a soft robot arm that bends and extends, but is torsionally stiff, requires a material with both soft and hard modes of deformation. Specifically, we need a material that is much stiffer in twisting than in bending and extension. Biological structures, such as plants, have been observed to evolve large differences in torsional and flexural stiffness as a method of adapting to dynamic environmental conditions \cite{Etnier2003-pn,Wolff-Vorbeck2019-zl,Wolff-Vorbeck2022-kp}. By altering their morphometric parameters (axial and polar second moment of inertia), many species of plants use the geometry of their stems and petioles to control the relative stiffness of their deformation modes. Similarly, creating soft robots with a large stiffness differential between flexure and torsion can be done by tuning their elastic response at the geometric level. This is possible using mechanical metamaterials, where mechanical behavior is a function of both mesoscale geometry and material properties \cite{Bertoldi2017-gb,Berger2017-nh}.

In this work, we present a robot made from a mechanical metamaterial that is stiff in torsion, yet soft under all other deformation modes. We call these metamaterials Torsionally Rigid Universal Couplings (TRUNCs). A single coupling acts as a soft constant-velocity joint (Fig.~S\ref{sup_fig:CV}). Multiple couplings can be connected in series and parallel to create nestable flex shafts that transmit independent concentric torques along complex curves. We present an arm composed of these joints that can interface with common tools such as socket drivers and wrenches while remaining soft and safe around humans. By bending and extending like an elephant's trunk, the arm increases its workspace size and gains additional movement dexterity. Our cable-driven arm achieves high movement repeatability for trajectories (0.4 mm and 0.1$\degree$) which outperforms the current state-of-the-art for soft multi-stage arms \cite{Guan2023-gj}. Finally, we demonstrate the real-world utility of our arm by having it install a lightbulb, assist a human with fastening bolts, and turn a shut-off valve. This work presents a robot arm that combines the safety and dexterity of soft robots with the utility of their rigid counterparts.    

\section*{RESULTS}

\begin{figure}[!h]
\makebox[\textwidth][c]{\includegraphics[width=\textwidth]{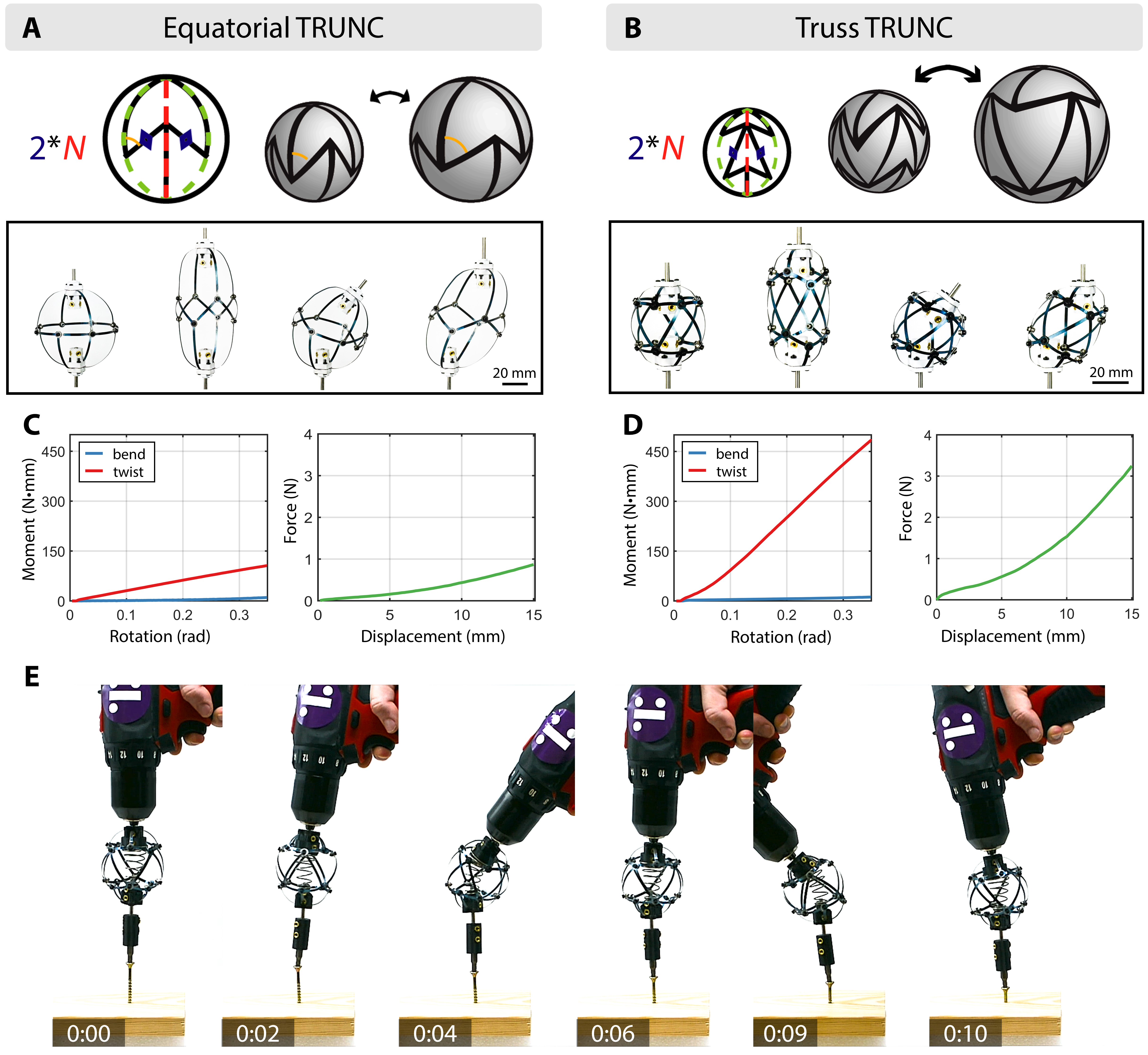}}
\caption{\textbf{{Fig. 2. \bf The two TRUNC variants and mechanical properties.} {\bf (A)} The equatorial TRUNC has a single band of joints along its equator {\bf (B)} and the truss TRUNC has two bands of joints on either side of the equator. This doubled structure allows for shear internal to the cell, which is disallowed in the equatorial structure. {\bf (C)} Moment-rotation in twist and bending modes and force-displacement in extension for the equatorial cell {\bf (D)} and truss cell demonstrates soft bending and extension compared to twist. {\bf (E)} A truss TRUNC performing a screwing operation at various bend angles, demonstrating torque transmission while bending. An internal conical spring provides restoring force.}}
\label{fig:Cell}
\end{figure}

\subsection*{Design and mechanical analysis}

Flexible structures for continuous torque transmission should be designed nestable and focused on achieving a high torsional stiffness without reducing the overall range of motion. The primary existing solutions, bellows \cite{Kim2024-nd} and flex shafts \cite{Tan2018-dj}, do not fully satisfy these performance metrics. Flex shafts are inextensible while bellows trade-off torsional rigidity with bending range of motion (See Supplementary Materials). We instead take inspiration from metamaterial design to demonstrate a family of materials to build soft robots. We introduce the Point Group Auxetics (PGA) to create Torsionally Rigid Universal Couplings (TRUNCs).

Conventional auxetics are often modeled as, and constructed of, rigid links that connect at rotational joints. Auxetics expand along a trajectory controlled by a single phase angle $\gamma$ that determines the expansion of the unit cell (identified for each cell with an arc in Fig.~\ref{fig:Cell}). As the phase angle increases, the cell expands in multiple directions up to a maximum phase angle $\gamma_{\text{max}}$ \cite{lipton632}. The symmetries of an auxetic are denoted here using orbifold notation. Orbifolds and their associated notation denote a symmetry group by its generating manifold \cite{Conway2002-fj}. 

The symmetry group building blocks for 3D structures are the 3D point groups. The 3D point groups are those that preserve a point through a discrete set of translations and rotations, and as such, they are also the tilings of the surface of a sphere. They break down into two subsets, 7 polyhedral groups and 7 axial point groups. The axial point groups preserve a single axis of rotation. The preservation of an axis generates two poles and an equator and orients the structure. Although the polyhedral groups are finite, the axial groups form infinite families where the rotational and mirror symmetries around the axis can vary from 2 to infinity \cite{conway2016symmetries}. 

Because the axial point group structures have a pole and equator, they can have three basic rotational responses: The poles can twist relative to each other to produce a net twist, the equator rotates relative to the polls, or there is no relative rotation. The \wallpapergroup{2}{*N} patterns, where N is any positive integer, maintain a structure along the equator of the sphere that connects to the poles, making them rigid to torsion.

We use these patterns to define a design space of TRUNCs, which encompasses their kinematic, geometric, and material parameters. The kinematic design space is defined by the tiling of a unit cell. The unit cell's geometry and properties are defined by the beam thickness and width (Fig. S\ref{sup_fig:tiling}A) relative to the neutral radius, and the Young's modulus ($E$) of the base material. These together determine the mechanical performance of a TRUNC.

We design TRUNCs as tilings of the auxetic double arrowhead structure seen in Figure~\ref{fig:Cell}. While there are $N$ tilings of the cell around the equator defined by the symmetry group \wallpapergroup{2}{*N}, there can also be $M$ tilings between the poles, where $M$ can be any number 2 or larger (Fig. S\ref{sup_fig:tiling}). As $M$ increases beyond 2, additional internal degrees of freedom are introduced. We therefore focus on two variations of the TRUNC joint that we call the equatorial ($M=2$) and truss ($M=3$) designs, shown in Fig.~\ref{fig:Cell} A and B respectively.

Both variants behave as a spherical mechanism on a sphere of non-fixed radius and act as constant velocity joints (see Supplementary Materials). The axial point group symmetry \wallpapergroup{2}{*N} reduces the easy modes of deformation to: 1) extension/compression along the axis when unit elements are deformed axisymmetrically and 2) bending along the axis produced by an antisymmetric deformation of unit elements. Twisting along the axis is a stiff mode, allowing transmission of rotation along composed axes. We choose a four-fold equatorial symmetry \wallpapergroup{}{*4} as it is the simplest family which produces pairs of cells for expansion and compression in each direction. The internal mode of the truss structure (M=3) allows shear internal to the cell, which can only be achieved through beam bending in the equatorial structure (Fig.~\ref{fig:Cell} A and B). The result is cells that can extend, bend, and shear while being rigid to torsion.

We characterized the selective compliance of both TRUNC variants using the ratio of torsional stiffness to flexural stiffness. The twist-bend ratio can be expressed as $GJ/EI$ using a specimen's elastic modulus ($E$), shear modulus ($G$), axial second moment of area ($I$), and polar second moment of area ($J$). We note that the twist-bend ratio is the inverse of the ratio used to study plants, which often aim to minimize torsional stiffness and maximize flexural rigidity \cite{Wolff-Vorbeck2019-zl,Naselli2023-kc}. We empirically characterize the flexural and torsional stiffness of both variants of the joints. The joints we characterized were made with spring steel (see Materials and Methods) and their stiffness values were found using linear regression on the moment-rotation data shown in Fig.~\ref{fig:Cell}~C and D.  Our results show that both variants are significantly stiffer in twisting than in bending. The equatorial TRUNC (Fig.~\ref{fig:Cell} left) has a twist-bend ratio of $11$ and the truss variant (Fig.~\ref{fig:Cell} right) has a ratio of $52$. In simulation, we investigated the fabrication design space parameters, defined by the material choice, cell diameter, and link width and thickness. We found that while the twist-bend ratio of both cells increased with a material's elastic modulus, the truss cell was less sensitive to changes between stiff materials (Fig. S\ref{fig:Ansys}). Furthermore, as the link's width and height increase relative to the cell's diameter, the twist-bend ratio increases, and the performance gap between both cell variants becomes smaller (see Supplementary Materials).  

Due to their large twist-bend ratios, TRUNCs can serve as highly flexible torque couplings. They can bend up to $45\degree$ from the neutral state while remaining $85.7\%$ efficient in torque and energy transmission (Fig S\ref{sup_fig:eff}A). This gives them the flexibility of rubber bellows while remaining much closer to the torsional rigidity of steel bellows, which can only bend up to $10\degree$. To illustrate the TRUNC's torsional rigidity, we connected a truss joint to a power drill and drove a screw into birch wood at various bend angles (Fig.~\ref{fig:Cell}G~Movie S1). A conical spring was added inside the cell to provide a restoring force without transmitting any torsional load, as its rotational degree of freedom is not rigidly constrained. This demonstrates how the TRUNC's large twist-bend ratio enables torque transmission through the hard mode while accommodating misalignment through the soft modes. To demonstrate the fatigue resistance of the cells we cycled 7 cells through over 6800 cycles with strains over 10\% without failure (see Supplementary Materials).

\subsection*{TRUNCs can be chained and nested}

\begin{figure}[p]
\makebox[\textwidth][c]{\includegraphics[width=\textwidth]{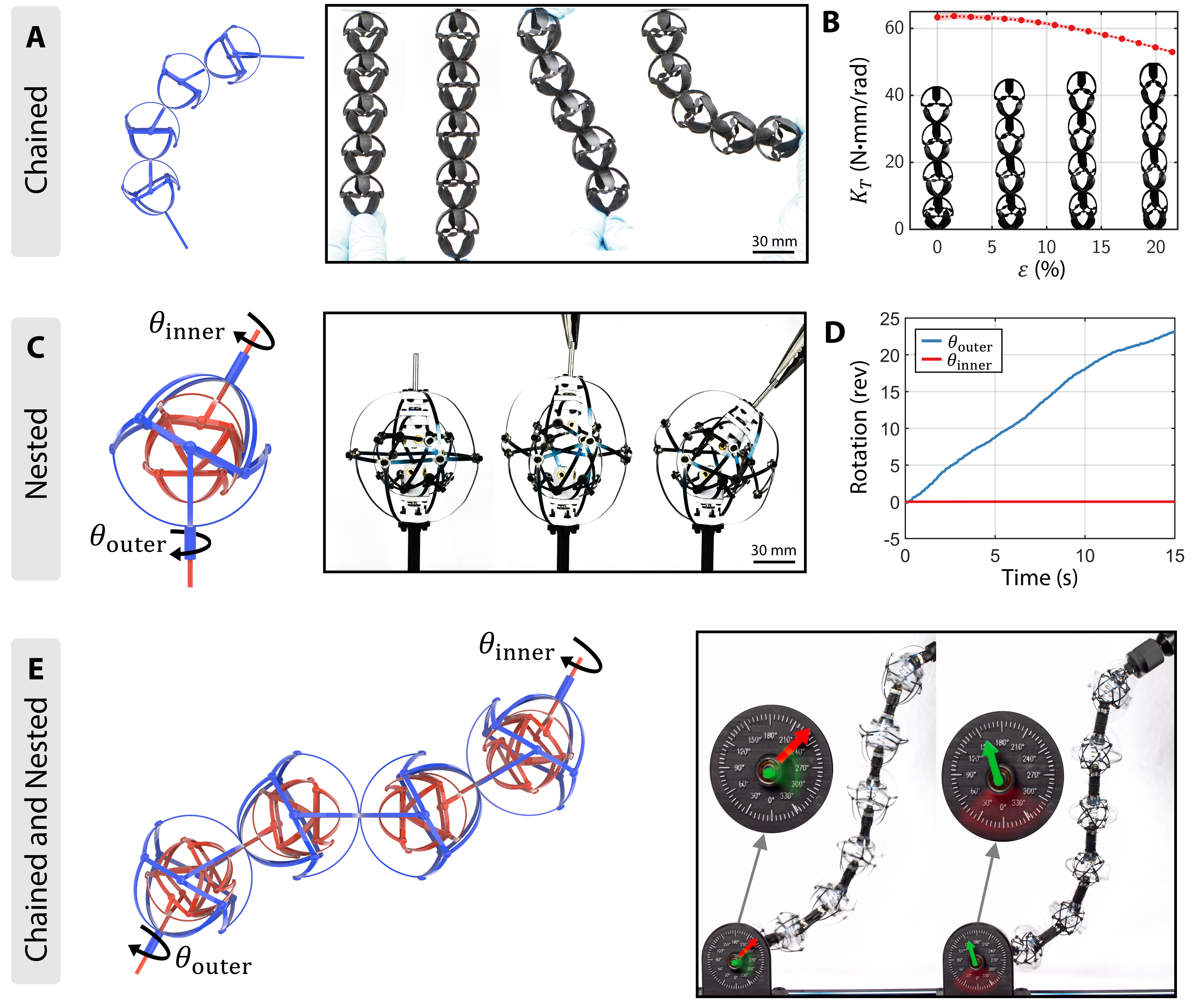}}
\caption{{Fig. 3. \bf Composition of TRUNCs.} {\bf (A)} Joints can be chained in series to create flex shafts that transmit torque while bending and extending. {\bf (B)} TRUNC flex shafts remain torsional rigid under extension. The shaded region represents the min and max bounds for $N=5$ trials. {\bf (C)} Joints can also be concentrically nested. Nested joints are coupled in bending and extension but  {\bf (D)} rotate independently. {\bf (E)} Nested joints can be chained to create flex shafts that transmit multiple torques. {\bf (F)} Each column of the flex shaft rotates separately, allowing multiple independent rotational degrees of freedom.}
\label{fig:Composition}
\end{figure}

The two TRUNC variants we previously introduced can be composed into larger metamaterials by connecting them in series or by concentrically nesting as shown in Fig.~\ref{fig:Composition}. When connected in series, fixed axis-to-axis, the TRUNCs act as a flexible shaft, conveying rotational motion and torque along a curved path. We show how a TRUNC made as a continuous material with living hinges can deform and transmit rotation. However, unlike traditional flex shafts, our design is extendable. When stretched by $20\%$ of its length, the TRUNC flex shaft maintained $83.6\%$ of its original torsional rigidity (Fig.~\ref{fig:Composition}B). Chained TRUNCs can mimic the extension of elephant trunks while maintaining their stiff torsional coupling properties.

To transmit multiple concentric torques, flexible torque couplings should nest efficiently inside each other. Because TRUNCS are based on spherical geometry, they can easily be nested as a series of concentric shells with ample room for clearance. This concentric composition is limited only by space, and can theoretically provide an infinite number of rotational degrees of freedom.  Nested TRUNCs bend and extend in unison but rotate independently (Fig.~\ref{fig:Composition}D). TRUNCs can be nested much more efficiently than bellows because more than 85\% of their volume is unoccupied, compared to 35\% or less in bellows couplings (see Supplementary Materials). When nested, a small cross-coupling effect exists between the cells due to friction within the bearings of the assembly (Fig. S3). This allows nested TRUNCS to transmit multiple independent torques through the same bend angle.  

When cells are nested and chained (Fig.~\ref{fig:Composition}E), they create a flex shaft that can transmit multiple torques independently along its length. To demonstrate this, we connected the flex shaft to a rotational gauge with two indicator dials. We then independently controlled the dials by switching the input torque between the inner and outer columns of the flex shaft (see Fig.~\ref{fig:Composition} and Movie S2). When the inner shaft is driven, the green dial moves, and the red remains still even as the arm deforms. When the outer shell is driven the inner green dial remains still. This demonstrates that the drill is only weakly connected to the dial indicator in terms of position, while the torsional connection remains strong. This allows us to build couplings that move freely through space while transmitting an arbitrary number of independent torques. Moving beyond soft materials to mechanical metamaterials enables us to build torque transmissions that rigidly couple input and output drives while remaining flexible in bending and extension.

\subsection*{A soft robotic arm composed of TRUNCs}

\begin{figure}[p]
\makebox[\textwidth][c]{\includegraphics[width=\textwidth]{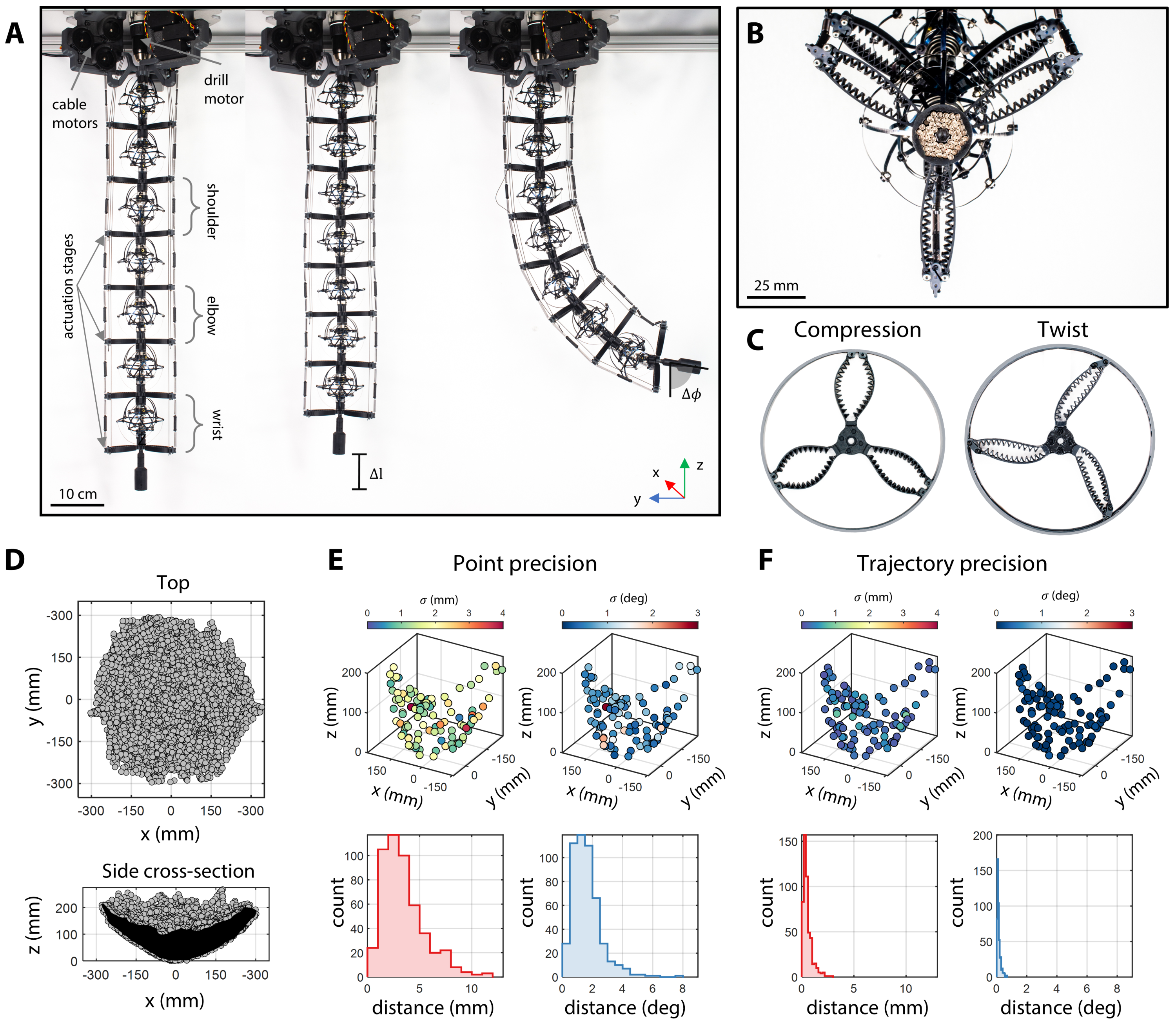}}
\caption{{Fig. 4. \bf Design and analysis of the TRUNC arm.} {\bf (A)} The body of the robotic arm consists of two nested flex shafts. Torque is transmitted to the end-effector using a drill motor mounted at the base. {\bf (B)} A head-on view of the arm with the socket driver end effector. {\bf (C)} The cables are connected via a structure that can bend and twist in-plane but resists out-of-plane bending. {\bf (D)} The workspace was empirically measured by sampling 18,300 poses. The bounding hull is concave and approximately symmetric about the $z$ axis. A cross-sectional view shows how the arm increases its workspace volume by extending along its length. {\bf (E)} The arm's positional and angular repeatability when visiting points in a random order (point precision)  and {\bf (F)} a fixed order (trajectory precision). Color maps represent the standard deviation (SD) across five trials.}
\label{fig:Arm}
\end{figure}

The nested TRUNC flex shafts we produce above enable us to create soft robots capable of continuous torque actuation. Our arm's design rigidly couples the end effector to a drill motor statically mounted at the base using a truss flex shaft. This layout leverages the benefits of direct motor drives while maintaining the robot's low inertia. The truss flex shaft is then nested within an equatorial shaft that guides the actuation tendons that control the arm (Fig.~\ref{fig:Arm}A). The cable actuation system uses nine servo motors mounted at the base. The nine cable lengths define the arm's configuration space $\mathcal{L}\in \mathbb{R}^9$ and encode all possible arm poses. Cables attach to the arm at three active joints we call the shoulder, elbow, and wrist. The remaining joints connected in between are passive. In its neutral state $\mathcal{L}_{\text{home}}$, the arm's cables are minimally pre-tensioned to a manually configured set point. An active joint can then be bent and compressed by varying the three corresponding cable lengths. This deformation is coupled to the passive joints above the active joint, allowing smooth bending and compression along each arm segment. We use springs to provide restoring forces that extend the arm as cable tension is released. The cables attach to the arm using spring structures that bend and twist in-plane, allowing them to remain compliant to contact forces (Fig. \ref{fig:Arm}C), but resist out-of-plane bending when actuated. Our arm design combines the advantages of soft bio-inspired arms with the utility of direct and continuous torque actuation.

Bending and extension enable the arm to position the end effector within a sizable workspace. We empirically measured the arm's reachable workspace by recording the end effector's position over a bounded sub-space of $\mathcal{L}$ consisting of 18,300 poses (see Materials and Methods). The boundaries of this subspace were found experimentally by driving the arm to extremal poses. The workspace was then reconstructed by fitting an alpha shape ($\alpha = 34.4$ mm) to the 3D point cloud, producing a concave hull with a volume of $\num[group-separator={,}]{18272}~\text{cm}^3$ (Fig. \ref{fig:Arm}D). The projection of the workspace onto the $xy$ plane approximates a circle with a diameter of ~600 mm or 84.5\% of the arm's neutral length ($l = 710$ mm). The cross-sectional view of the $xz$ projection shows the workspace is symmetric about the $z$ axis and illustrates how the extension degree of freedom increases the total volume. We note that the workspace is thickest near the origin and becomes thinner near the edges. This thinning effect occurs because the arm undergoes additional compression while bending, which reduces its range of motion. From our workspace analysis, we found that the arm can compress up to $\Delta l = 94.3$~mm, or $13.3\%$ of its length in the $z$ direction, and tilt the end effector up to $83.9 \degree$ from the ground plane.

Within this recorded workspace, we characterized the positional and angular repeatability of the arm's movements. Due to non-linear and hysteretic effects, a soft robot's state at any time is influenced by its history of previous states \cite{Thuruthel2019-nt}. Thus, we used separate tests to measure repeatability for point positioning and trajectory following. We generated the test trajectory by randomly selecting 100 configurations and ordering them using K-nearest neighbors (KNN) as a greedy solution to the traveling salesman problem. The point repeatability test used the same points but randomized the order at the start of each trial. Before each trial, the arm cycled five times between its neutral and fully compressed state to reset any hysteretic effects from previous tests. For each test, the arm visited the 100 points over five trials and we calculated the translational and angular distance of each data point from its cluster's mean. For point positioning, we found the standard deviation (SD) of the residual distances to be $2.1$ mm and $0.1 \degree$ (Fig. \ref{fig:Arm}E). When tracking a trajectory, the SD decreased to $0.4$ mm and $0.1 \degree$ (Fig. \ref{fig:Arm}F). With millimeter-level trajectory repeatability, our arm outperforms state-of-the-art multi-stage soft arms on trajectory precision\cite{Guan2023-gj}.

\subsection*{Experimental demonstration}

\begin{figure}[p]
\makebox[\textwidth][c]{\includegraphics[width=\textwidth]{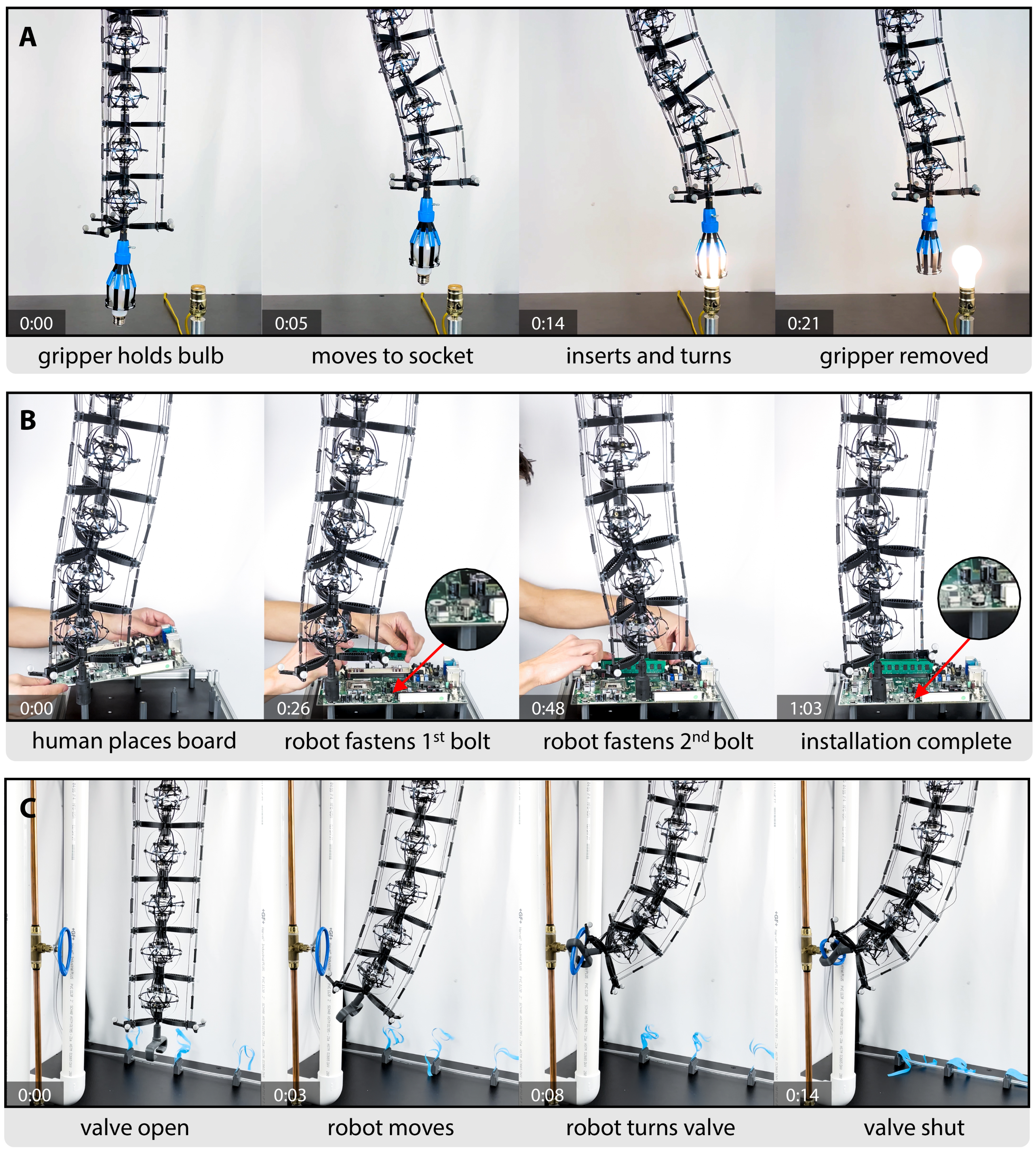}}
\caption{Fig. 5. {\bf Demonstrating applications for the TRUNC arm.} {\bf (A)} Delicately installing a light bulb into a lamp. The arm positions the bulb and then rotates it until the bulb turns on. {\bf (B)} Safe human-robot collaboration for motherboard installation. The robot secured the board to the frame by fastening bolts while the human installed RAM modules. {\bf (C)} Manipulating a valve positioned between two PVC pipes to shut off airflow.}
\label{fig:Demo}
\end{figure}

Continuous torque actuation can enable soft robots to perform a broader range of real-world tasks than their completely compliant counterparts. We trained a neural network to learn the arm's inverse kinematics (see Materials and Methods) and demonstrated the TRUNC arm installing a light bulb, fastening bolts for a motherboard, and shutting off a valve. For these experiments, we measured the position of target objects (see Materials and Methods) and then used the learned inverse kinematics model to generate each trajectory. In the first experimental demonstration, the arm installed a light bulb into a lamp (Fig.~\ref{fig:Demo}A and Movie S3). Our end effector was a commercially available universal light bulb changer that grips the bulb using rubber-coated steel. The arm first positioned the bulb above the socket before lowering it into place. Compliance enables the arm to passively align itself for peg-in-hole insertion tasks. The arm leveraged this by extending past the targeted $z$ height to increase the insertion force, helping the bulb's threads engage once the bulb starts rotating. Once positioned within the socket, the bulb was rotated by $2~\frac{1}{2}$ revolutions which completed the circuit and turned it on. The arm then disengaged with the bulb by pulling the end effector up before returning to the home position. This demonstrates that even using simple motion planning, the arm can carefully handle and manipulate delicate objects and perform complex fastening operations.

In the next experiment, we used the TRUNC arm to work alongside a human and install a motherboard on a mounting frame (Fig.~\ref{fig:Demo}B and Movie S4). The assembly began with the human placing the board on the stand-offs and pre-fastening the M6 bolts by two turns each while the arm waited nearby. Once the board was placed, the arm received a command to begin fastening bolts that secured the board to an aluminum frame. As the arm fastened bolts, the human worked alongside it by installing RAM modules. The arm was programmed to position the socket above each fastener before plunging in the $-z$ direction. The end effector's universal socket adapter (Fig.~\ref{fig:Arm}B) has spring-loaded pins that decouple the need for simultaneous extension and rotation during thread fastening operations. This enabled the arm to hold its position while transmitting torque to fasten each bolt by $7 \frac{3}{4}$ revolutions. Automating this thread fastening sub-routine allowed the human co-worker to continue working safely in parallel with the robot.   

Finally, we showed how the arm can manipulate objects within constrained spaces. This demonstration simulated a leak scenario where the arm had to reach and turn a valve positioned between two pipes (Fig.~\ref{fig:Demo}C and Movie S5). We used a 3D-printed wrench end effector that passively grips and locks onto the valve once it begins turning. The arm's continuous end effector rotation simplifies the task as the gripper does not need to disengage and reengage while operating in a tight space. The arm combined bending with linear translation to reach the valve and slide the end effector over it. A smooth motion transition in orientation was achieved using spherical linear interpolation between the home and target quaternions. Once engaged with the valve, the arm applied $3~\frac{1}{2}$ clockwise rotations which successfully shut off airflow as indicated by the outlet's blue streamers. While actuating the valve, the arm passively deformed to compensate for misalignment between the wrench and valve's center of rotation.

\section*{DISCUSSION}

We presented the TRUNC arm, a flexible soft robot that possesses the utility of continuous torque actuation. The arm's design is based on a pair of soft extendable constant-velocity joints capable of continuous torque actuation. We introduced and characterized two variants of the joint based on the same symmetry group. The joints achieved a wide range of motion in compression and  bending. The selective compliance architected into their design, with a low axial and bending stiffness but high torsional rigidity, allows TRUNCs to be coupled with motors to create soft actuators that can transmit torque.

We showed how composing TRUNCs enables them to transmit multiple torques over longer distances. The first composition method connects joints in series to create flexible shafts capable of bending and extending, while nesting joints inside each other enabled concentric torque transmission. Combining both methods, we created a nested flex shaft with two rotational degrees of freedom about the central axis that we independently actuated in Fig. \ref{fig:Composition}E.

We applied these principles to create a soft arm with continuous torque actuation. The cable-driven arm transmits torque to its end effector using a flexible TRUNC shaft connected to a motor mounted at its base. We demonstrated that the arm can compress up to  $13.3\%$ of its length and tilt the end effector up to $83.9\degree$ from the ground plane. The arm can position the end effector within a large workspace by bending, compressing, and extending, and moves with high motion repeatability along trajectories (0.4 mm and 0.1$\degree$). This enables us to perform many real-world tasks that were previously inaccessible to soft robots.

We chose three tasks to demonstrate the versatility of the TRUNC arm using different end effectors in different environments. The first task, installing a light bulb, demonstrated that the arm has the mobility to extend and twist while manipulating delicate objects and to passively align itself when fastening threads. In the second, the arm works safely around a human to install electronic components. In the third task, the soft arm was able to navigate a close environment to shut off a valve by rotating it continuously over multiple turns. Using selective compliance, the arm combines flexibility with robust manipulation. This work bridges the gap between soft and rigid robots by combining the safety of compliant structures with the utility of rigid and continuous torque transmission.

\section*{MATERIALS AND METHODS}

\subsection*{Cell and arm fabrication}
The monolithic flexure TRUNCs were printed on a Carbon M1 printer using FPU 50 resin. Spring steel TRUNCs were assembled from laser-cut 0.01" 1095 hardened spring steel. Technical drawings are provided in Supplementary Materials. Spring steel parts were sanded around the joint locations to expose the bare metal, removing defects, burs, and surface coatings. Steel links were manually bent using a spherical mold with a 56 mm diameter for the truss cells and an 88 mm diameter mold for equatorial cells. The links were then assembled using M2 screws and nylon lock nuts. Chained cells were connected using a 4 mm steel rod press-fit into a bearing (McMaster part ID: 6655K47) which was fastened to the cell with a 3D printed connector. Heat-set threaded inserts were installed to fix the coupling shafts to the cells. Nesting cells, were offset with a thrust bearing (McMaster part ID: 6655K47) between the contacting surfaces to reduce friction. 

We used nine Hitec HS-785HB winch servos for the cable actuation system. The home configuration ($\mathcal{L}_{\text{home}}$) was found during the arm's setup by controlling each motor until its cable's slack was removed. The cable attachment triads (Fig. \ref{fig:Arm}C) were fabricated on a Prusa MK4 extrusion 3D printer in PLA+. The triad structures route the cables through an embedded Polytetrafluoroethylene (PTFE) lining to reduce friction. Actuating cables are 0.75 mm synthetic braided cable. A conical spring ($K=1.22$ N/mm) is mounted inside each cell with a free length longer than the inner sphere diameter so that the cells are pre-tensioned, providing a restoring force for the arm. The arm also has six extension springs ($K=0.07$ N/mm) for each cell that provide restorative forces when bending. Torque was transmitted to the end effector using an 18V DC Milwaukee drill motor (connected to $2.15$ V power supply) and was toggled using a relay module. The motors and arm were mounted to a 0.25" aluminum base plate that was fixtured to the motion capture booth.

\subsection*{Kinematic characterization using rotational encoders}

Kinematic tests for the joints (Fig. \ref{fig:Composition}D and Fig. S\ref{sup_fig:CV}) were conducted using two YUMO E6B2-CWZ3E rotational encoders connected to a National Instruments (NI) USB-6009 Data Acquisition (DAQ) unit.  The input and output shafts were connected to separate encoders mounted on a steel fixture plate and a 12V DC motor was then used to drive the input shaft.  Data was sampled at a rate of 10,000 Hz and the quadrature signals were read and decoded in MATLAB.

\subsection*{Mechanical characterization}

The tensile, torsional, and flexural tests were performed using an Instron CCP122055 biaxial load cell $\pm450$ N and $\pm5$ $\text{N}\!\cdot\!\text{m}$. For the spring steel TRUNCs, we applied WD-40 to the spring to reduce friction. In our bending and torsion tests, we rotated the cell by $-5\degree$ before returning to the origin to reduce backlash effects. Data was collected at intervals of $1\degree$ with an angular velocity of $1$ rev/min. In our extension tests, we pre-compressed the cell to $-2$ mm before returning to the origin. Data was then collected at intervals of $0.2$ mm with a linear velocity of $0.5$ mm/s. 

\subsection*{Motion capture setup}

We captured the ground truth pose of the arm using an OptiTrack motion capture system with eight Flex 13 cameras and streamed the data to MATLAB using NatNet. The workspace dataset (Fig. \ref{fig:Arm}D) and trajectory tests (Fig. \ref{fig:Model}) tracked the end effector using the socket tool with four markers affixed at the base. Because the inverse model was trained on data collected with the socket end effector, we measured a coordinate transformation for each new tool to compensate for geometric differences when generating new trajectories. For our experimental demos (Fig. \ref{fig:Demo}) we measured the position of the target objects (bolts, socket, and valve) using 3D printed mounts for motion capture markers prior to testing.

\subsection*{Neural network training}

We trained our inverse kinematic model on the workspace dataset (Fig. \ref{fig:Arm}D), consisting of 18,300 pairs of configuration states and end effector poses. Randomly sampling the configuration space of cable-driven results in ill-defined configurations where the arm loses actuation degrees of freedom due to slack or `dead-bands' in the cables. Our learning approach sampled $\mathcal{L}$ using an approximate kinematic model (see Supplementary Materials) and refined these configurations during the data collection process by slowly tensioning each cable until the end effector moved by more than $3$ mm or $1 \degree$. Additionally, we enforced a bijective mapping to $SE(3)$ by virtually coupling the shoulder and elbow rotations (see Supplementary Materials). The dataset was normalized using min-max normalization and then split into training (80\%) and validation (20\%) sets.

The DNN was implemented in PyTorch and consists of three fully connected layers. The input layer size (7 neurons) corresponds to the end effector's position ($x,y,z$) and orientation represented as a quaternion ($q_w,q_x,q_y,q_z$). There are two hidden layers with 1600 neurons each and ReLU activation functions to introduce non-linearity. Finally, the output layer has 9 neurons corresponding to the dimensions of $\mathcal{L}$. For our loss function, we used the Mean Squared Error (MSE) between the predicted and actual arm configurations. When training, we used the Adam optimizer with a learning rate of 0.001, momentum of 0.9, and no weight decay. The learning rate was annealed during using exponential decay with $\lambda = 0.9$. We used a batch size of 16 to balance computational efficiency and the granularity of learning. The model was trained for 50 epochs which took 120.6 seconds using a cloud-based GPU (Nvidia Tesla v100). The final mean training error was $1.04$ mm per cable while the mean validation error was $1.11$ mm per cable. When performing inference to generate trajectories, the model was run on the computer's CPU (12th Gen Intel i7-12700 2.10 GHz) and took on average 0.368 ms per waypoint. Trajectory following results are provided in Supplementary Materials.



\section*{ACKNOWLEDGMENTS}

\textbf{Funding}: This work was funded by the National Science Foundation grant number 2212049, a grant from The Murdock Charitable Trust, and the Office of Naval Research grant number 304220. \textbf{Author Contributions:} M. C. and J.I.L designed the cell; M.C., J.F.K.,  and J.G. designed the robot; M.C., J.F.K., A.G., and D.R., modeled the robot; M.C., J.F.K., J.G., and J.F.A., tested the cell; M.C., J.F.K., J.G., and J.F.A. tested the robot; M.C., J.F.K., J.F.A., D.R., and J.I.L. wrote the paper; J.I.L provided funding.  \textbf{Competing Interests:} We have filed a patent: US Patent App. 17/759,826. \textbf{Data and materials availability:} All data and materials will be publicly available at https://github.com/TransformativeRoboticsLab/TRUNC.

\section*{SUPPLEMENTARY MATERIALS}

Supplementary Text\\
Figs. S1 to S7\\
Movies S1 to S5

\renewcommand\refname{REFERENCES AND NOTES}

\bibliography{scibib}

\bibliographystyle{Science_forrobotics}


\clearpage

\pagenumbering{gobble}

\begin{center}
    \vspace*{\stretch{0.25}} 
    \LARGE{Supplementary Materials for}
    \\
    \vspace{0.5cm}
    \textbf{\Large{Bridging Hard and Soft: Mechanical Metamaterials
Enable Rigid Torque Transmission in Soft Robots}} 
    \vspace{1cm} 
    
    \large Molly Carton \it{et al.} 
    \vspace*{\stretch{0.5}} 
\end{center}

\noindent \textbf{The PDF file includes:} \\
Supplementary Text \\
Figs. S1 to S9 \\

\noindent \textbf{Other Supplementary Material for this manuscript includes the following:} \\
Movies S1 to S5

\newpage

\section*{Supplementary Text}

\setcounter{figure}{0} 

\subsection*{Constant-velocity kinematics}

\begin{figure}[!h]
\makebox[\textwidth][c]{\includegraphics[width=\textwidth]{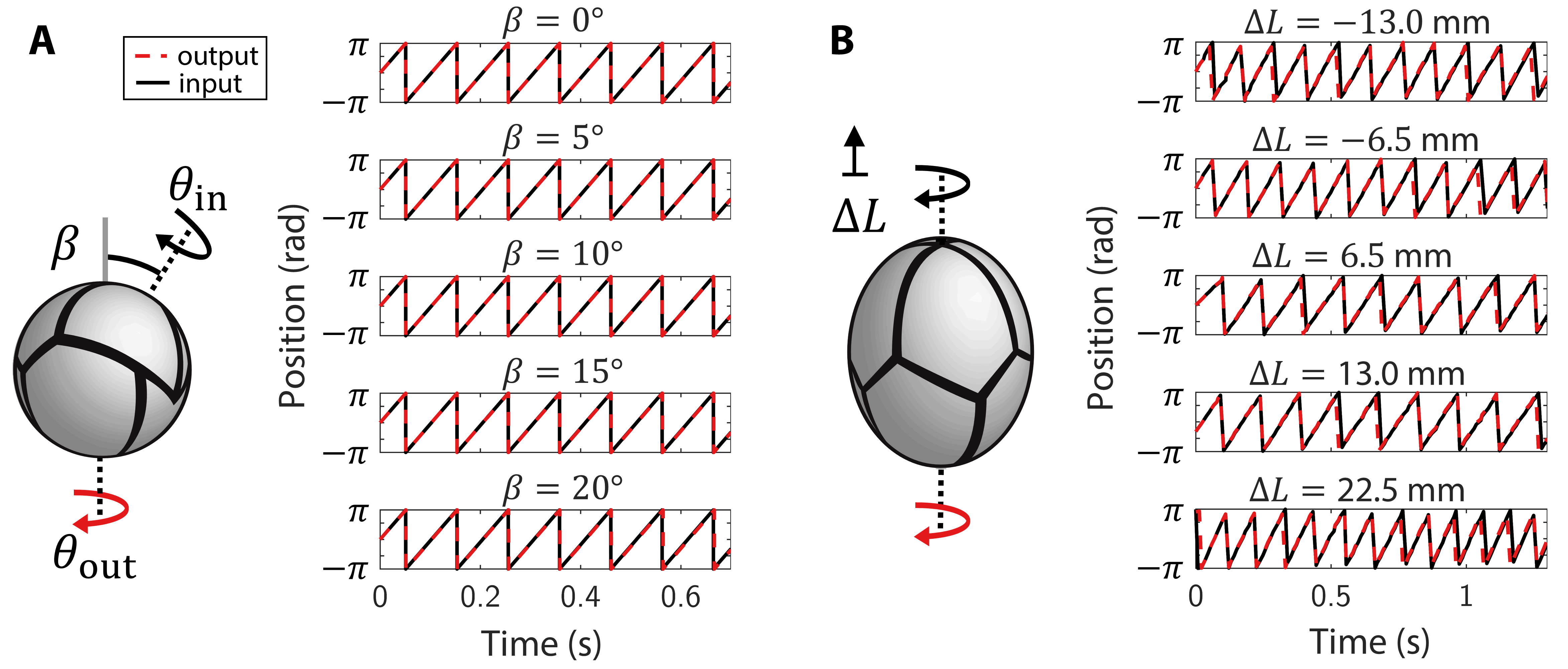}}
\caption{{Fig. S2. \bf TRUNCs are constant-velocity joints.} TRUNCs maintain a constant angular velocity between input and output shafts when {\bf (A)} bending and {\bf (B)} extending. Input and output signals are plotted in radians (modulo 2$\pi$).}
\label{sup_fig:CV}
\end{figure}

Constant velocity joints are kinematic couplings that maintain equal speeds between input and output shafts while bending. We characterize the spherical element as a constant-velocity joint by performing a phase-matching test (Fig.~\ref{sup_fig:CV}). The single spring-steel joints were fixtured with varying (A) axial bend angle and (B) extension and driven using a DC motor, while angular position encoders measured the input and output position and velocity. These results demonstrate a close phase-match between the driven input and output shafts.

\subsection*{Design space analysis}

\begin{figure}[!h]
\makebox[\textwidth][c]{\includegraphics[width=\textwidth]{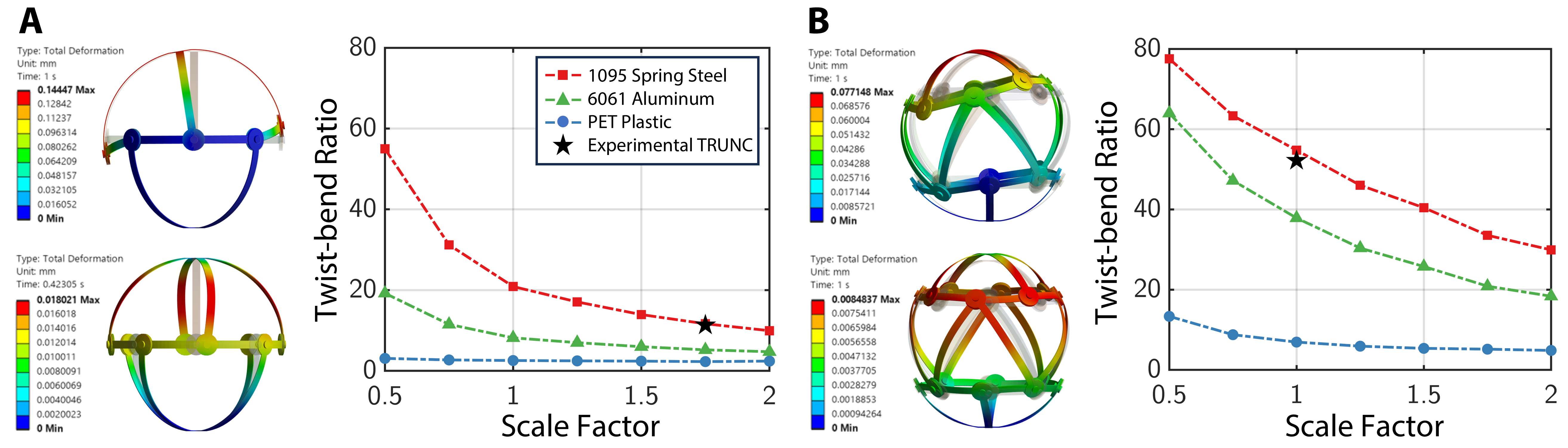}}
\caption{{Fig. S3. \bf Design space analysis of the TRUNCs.} The twist-bend ratio, calculated from simulation using Ansys Static Structural, versus size scale for the (A) equatorial and (B) truss TRUNCs, comparing varying material models against experimental results.} 
\label{fig:Ansys}
\end{figure}

We used Finite Element Analysis (FEA) to understand how size scale and material selection impact a TRUNC's twist-bend ratio (Fig. S\ref{fig:Ansys}). In our analysis, we varied the TRUNC's diameter while holding the width and thickness of each link constant. The diameters were normalized using the truss joints in the arm ($D = 56$ mm). Models were loaded into Ansys Static Structural Analysis and pin joints were defined as revolute connections. Friction in the joints was approximated using a torsional stiffness of 0.05 $\text{N}\!\cdot\!\text{mm}/^\circ$. Large deflections and weak springs were enabled and the mesh's element order was set to quadratic to improve the solver's converge. The equatorial TRUNC was meshed with approximately 7500 elements, while the truss TRUNC was meshed with approximately 14000 elements. The twist-bend ratios were calculated by applying a 1.5 $\text{N}\!\cdot\!\text{mm}$ twisting moment followed by a 0.15 $\text{N}\!\cdot\!\text{mm}$ bending moment to the cell's top surface while the bottom was fixed. Across three trials, we assigned the link material as Polyethylene terephthalate (PET), 6061 Aluminum, and 1095 Spring Steel (hardened and tempered) with elastic moduli of 2.9 (GPa), 69 (GPa), and 211 (GPa) respectively. We also simulated Neoprene Rubber but excluded these results from the analysis because the structure deformed without the links rotating about the pins due to insufficient rigidity. The pins were always assigned as Structural Steel with a modulus of 200 (GPa). Loading moments were normalized based on the material's elastic modulus. 

Our simulation results show increasing the TRUNC's material stiffness resulted in a higher twist-bend ratio for both the equatorial (A) and truss (B) variants. Furthermore, the truss cell was less sensitive to changes in modulus between stiffer materials. Moving from Aluminum to Steel, a 190\% increase in modulus, yielded a $124\%$ and $44\%$ increase in the twist-bend ratios of the equatorial and truss cells respectively. Additionally, we found that the twist-bend ratio is inversely related to the scaling factor. Thus, scaling the link's width and thickness relative to cell diameter can increase the twist-bend ratio. Doubling the relative thickness and width resulted in the twist-bend ratios increasing by $472\%$ and $140\%$ for the equatorial and truss cells respectively. These results suggest that a future design optimization of the link parameters, relative to the cell size, could substantially increase the TRUNC's selective compliance.

\subsection*{Monolithic TRUNCs}

\begin{figure}[!h]
\makebox[\textwidth][c]{\includegraphics[width=\textwidth]{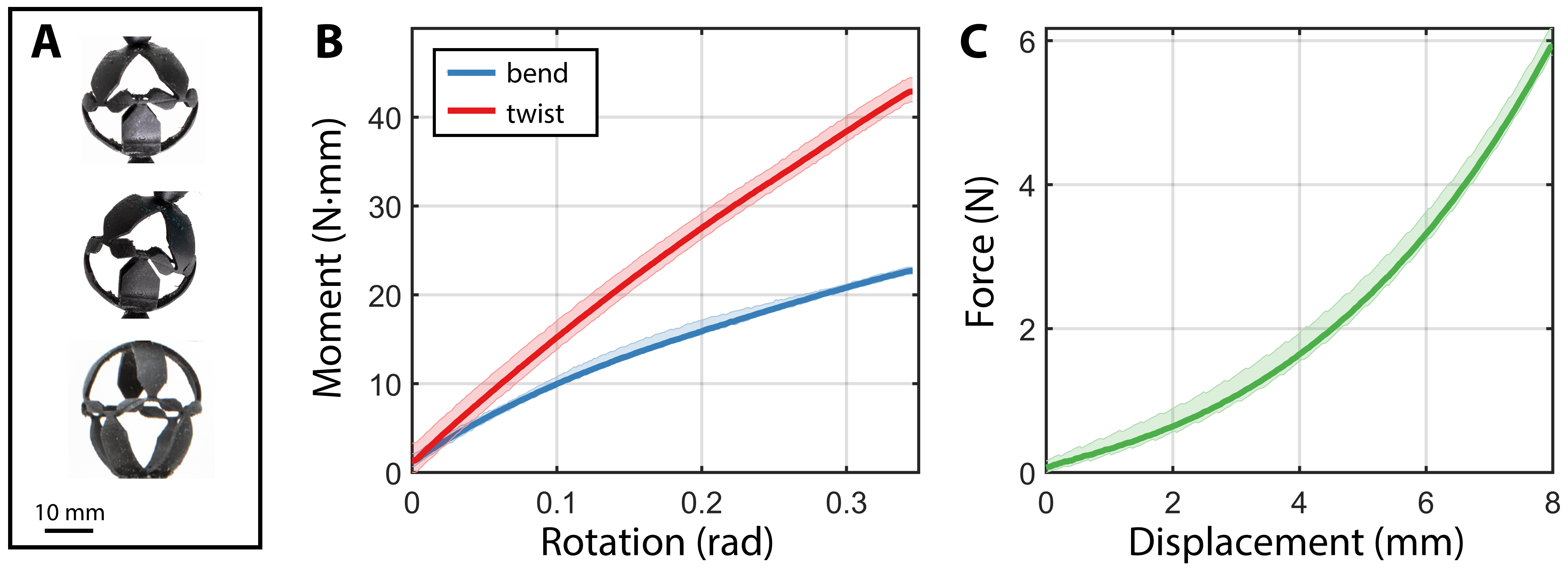}}
\caption{{Fig. S5. \bf Monolithic TRUNC characterization.} {\bf (A)} A monolithic equatorial TRUNC printed from FPU 50 demonstrating polar bending mode and extension. We characterized the mechanical properties of a single printed cell for {\bf (B)} torsional and flexural stiffness {\bf (C)} and axial stiffness. Shaded error bars represent the min and max values for five trials.} 
\label{fig:Printed}
\end{figure}

Monolithic (3D-printed flexure-based) modules were designed using Autodesk Fusion 360. Joint features and aspect ratios were determined empirically, to achieve approximate pin joint behavior. Part performance was evaluated using an Instron 68SC2 test frame with a 450 N/5 $\text{N}\!\cdot\!\text{m}$ torsion-tension load cell. Rotational stiffness and load/displacement for a single printed joint are shown in Figure~\ref{fig:Printed}. In Figure~\ref{fig:Printed} A, we show the bending and extension of the monolithic joint. Part B demonstrates the joint has a twist-bend ratio of 2. These printed flexure mechanisms demonstrate the material-agnostic nature of this structure. While the idealness of the mechanism is dependent on geometric properties, with appropriate translation to flexure joints it is possible to realize these mechanisms in a single-material monolithic structure. We use this to demonstrate that stiffness remains uncoupled from extension despite the jointed flexure of the constituent material.

\subsection*{Cross-coupling of nested joints}

\begin{figure}[!h]
\makebox[\textwidth][c]{\includegraphics[width=\textwidth]{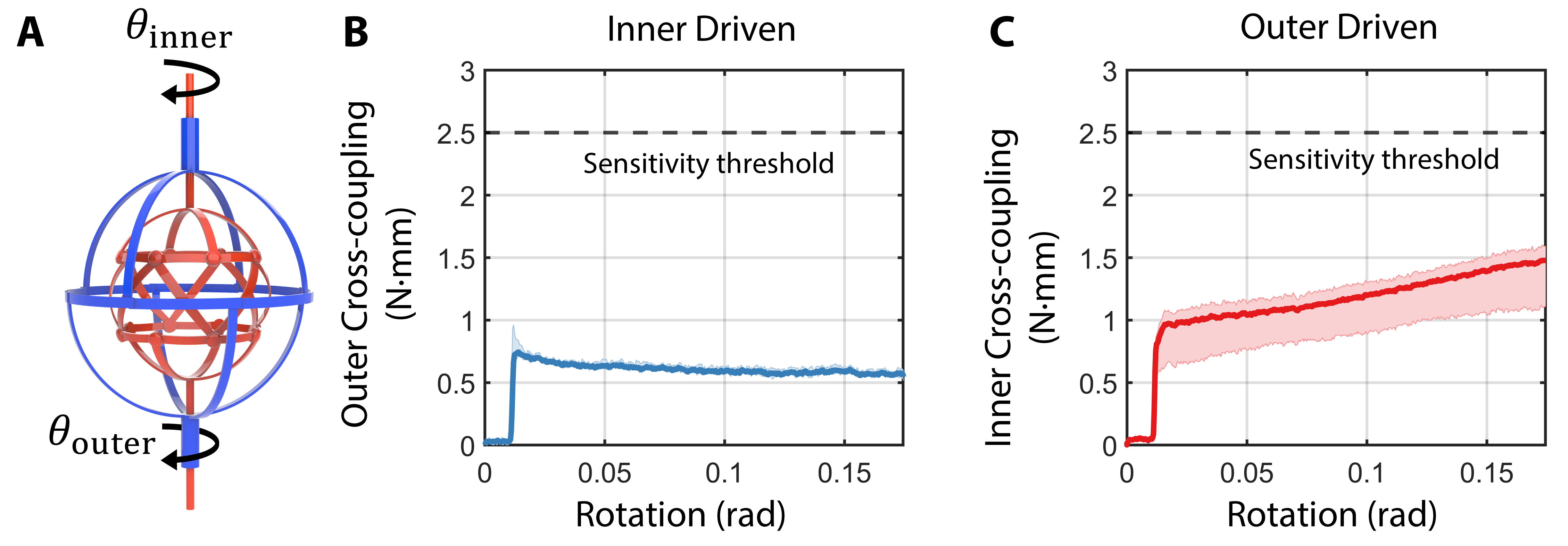}}
\caption{{Fig. S7. \bf Cross-coupling analysis of nested TRUNCs.} Nested joints experience a small amount of rotational cross-coupling due to friction. We measured the cross-coupling torque experienced when {\bf (A)} driving the inner truss joint and {\bf (B)} the outer equatorial joint. Shaded error bars represent the min and max values for five trials.} 
\label{fig:Cross-coupling}
\end{figure}

Cross-coupling between nested joints occurs due to friction in the system. We measured this effect when driving the inner and outer couplings for the nested spring-steel TRUNC (Fig. \ref{fig:Composition}C). For these tests, we connected the driven side to the Instron's upper grippers which rotate the specimen, while the passive coupling was connected to the load cell via the lower grippers. The driven coupling was rotated at $1$ rev/min for $10 \degree$ (0.174 rad). We measured cross-coupling when driving the inner (Fig. S\ref{fig:Cross-coupling}A) and outer (Fig. S\ref{fig:Cross-coupling}B) TRUNCs. Each experiment was repeated $n=5$ times to obtain uncertainty bounds. We found that in both experiments, the cross-coupling torque remained significantly below the Instron's calibrated sensitivity of $0.05$ \% of the maximum rated load or $2.5$ $\text{N}\!\cdot\!\text{mm}.$ This represents $0.51 \%$ and $2.38 \%$ of the torque capacity of the truss and equatorial cells respectively.

\subsection*{Performance Comparison}

\begin{figure}[!h]
\makebox[\textwidth][c]{\includegraphics[width=\textwidth]{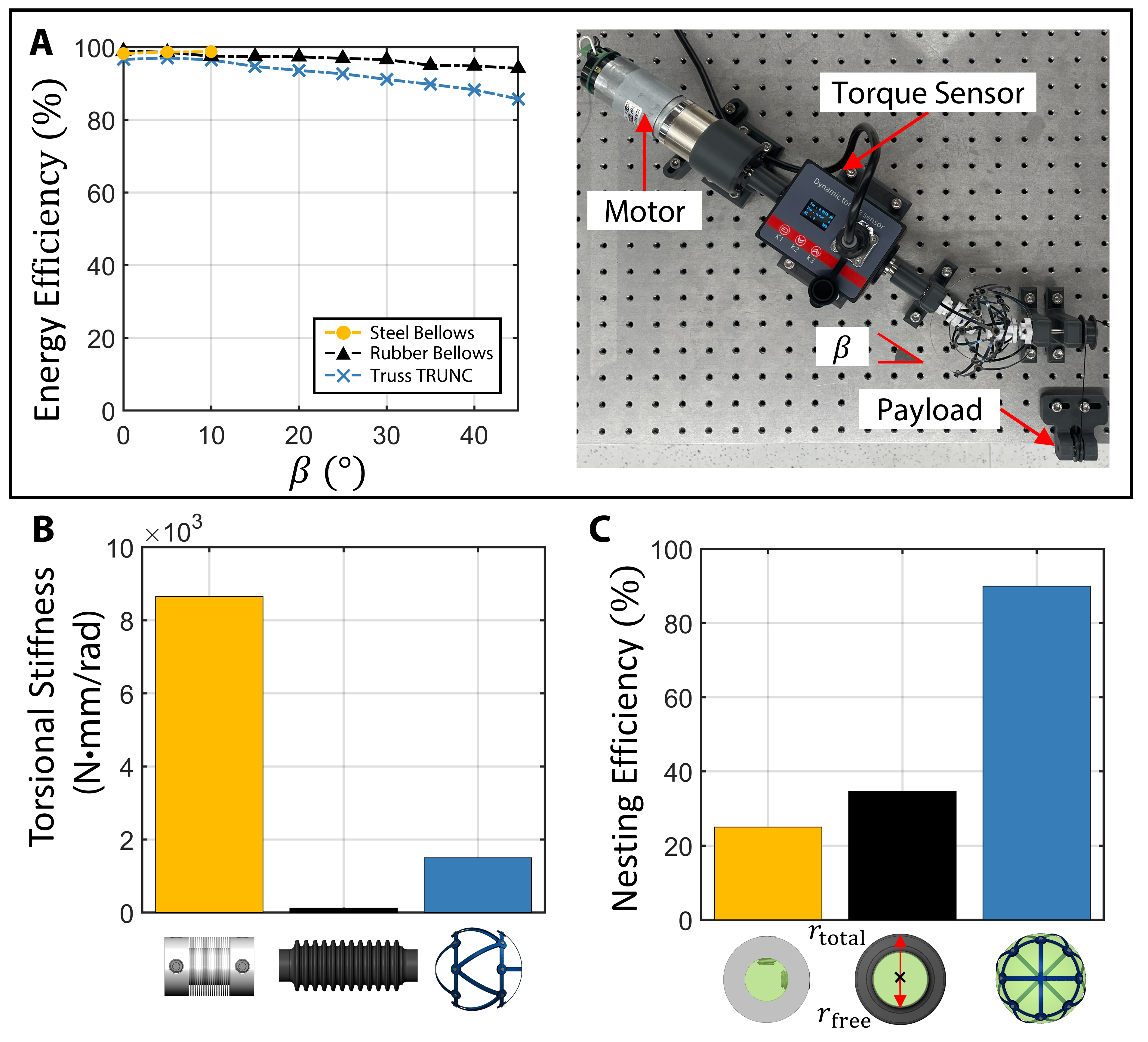}}
\caption{{Fig. S6. \bf Performance comparison of TRUNCs and bellows.} (\textbf{A}) We characterized the energy efficiency of the nested truss TRUNC as a function of its bend angle $\beta$. As a point of comparison, the same experiment was performed with off-the-shelf rubber and steel bellows couplings demonstrating that the TRUNC achieves an energy efficiency close to that of the rubber couplings for the same large range of motion.(\textbf{B}) Comparison of torsional rigidity for off-the-shelf steel and rubber couplings and truss cell. The TRUNC cell achieves a torsional rigidity between that of the steel and rubber couplings while maintaining the wide angle range of the rubber coupling. (\textbf{C}) Comparison of cross-sectional nesting efficiency. The cross-sectional radii ($r_{\text{free}}$ and $r_{\text{total}}$) were matched to the CAD models. TRUNCs can nest more efficiently within each other compared to bellows because they lie on the surface of a sphere, whereas bellows occupy a cylindrical volume.} 
\label{sup_fig:eff}
\end{figure}

We compared the TRUNCs to commercially available steel and rubber bellows couplings which can accommodate misalignment when transmitting torque. For the bellows, we purchased a 303 Stainless Steel and Nickel Alloy bellows coupling (McMaster part ID: 59925K88) and a Neoprene Rubber bellows (McMaster part ID: 5298K23). We characterized the energy efficiency $\eta = \frac{P_{\mathrm{out}}}{P_{\mathrm{in}}}$ of each coupling as a function of their bend angle $\beta$ (Fig. S\ref{sup_fig:eff}A). For each bend angle, we conducted two tests to measure $P_{\mathrm{input}}$ and $P_{\mathrm{output}}$ using an ATO-TQS-DYN-200 (1 $\text{N}\!\cdot\!\text{m}$) dynamic torque sensor. Input torque was supplied using a 12V DC motor. Before testing, WD-40 Specialist White Lithium Grease was applied to the TRUNC's joints. A hanging mass was attached to the output to create a 0.1 $\text{N}\!\cdot\!\text{mm}$ load. Up to a small bend angle ($\beta = 10^\circ$), the steel bellows was the most efficient ($\eta = 98.8\%$) followed closely by the rubber bellows ($\eta = 97.5\%$), and TRUNC ($\eta = 96.5\%$). The rubber bellows and TRUNC could both bend up to $\beta = 45^\circ$ and achieved efficiencies of $\eta = 94.2\%$ and $\eta = 85.7\%$ respectively. In future work, the TRUNC's efficiency at larger bend angles could be potentially improved by switching the joints, currently made using fully threaded M2 bolts, to smoother shoulder screws.

While the rubber bellows bent almost five times further than steel bellows, its torsional stiffness was much lower. TRUNC's achieved the same large bending range of motion as the rubber bellows while remaining closer to the torsional stiffness of steel bellows (Fig. S\ref{sup_fig:eff}B). The torsional stiffness for the bellows was calculated using the same mechanical characterization techniques discussed in Materials and Methods for the TRUNCs. Furthermore, TRUNCs can be more efficiently nested inside each other than bellows. As a scale-invariant metric for nesting efficiency, we used the ratio of unoccupied volume (shown in green in Fig. S\ref{sup_fig:eff}C) to total volume. For hollow cylinders with length $l$, total radius $r_\text{total}$, and unoccupied radius $r_\text{free}$, the nesting efficiency is:
\[
\mathcal{N}_\text{cylinder} = \frac{l \: \frac{\pi}{4} \: r_{\text{free}}^2}{l \: \frac{\pi}{4} \: r_{\text{total}}^2} = \frac{r_{\text{free}}^2}{r_{\text{total}}^2} 
\]
For hollow spheres, the nesting efficiency is:
\[
\mathcal{N}_\text{sphere} = \frac{4 \: \pi \: r_{\text{free}}^3}{4 \: \pi \: r_{\text{total}}^3} = \frac{r_{\text{free}}^3}{r_{\text{total}}^3} 
\]
Because TRUNCs lie on the surface of a sphere, they occupy much less volume, with approximately $85\%$ of the space being free. On the other hand, bellows must occupy a thick-shell volume inside of a cylinder to gain flexibility and have less than $35\%$ unoccupied volume.

\subsection*{Cyclic loading and durability}

\begin{figure}[!h]
\makebox[\textwidth][c]{\includegraphics[width=\textwidth]{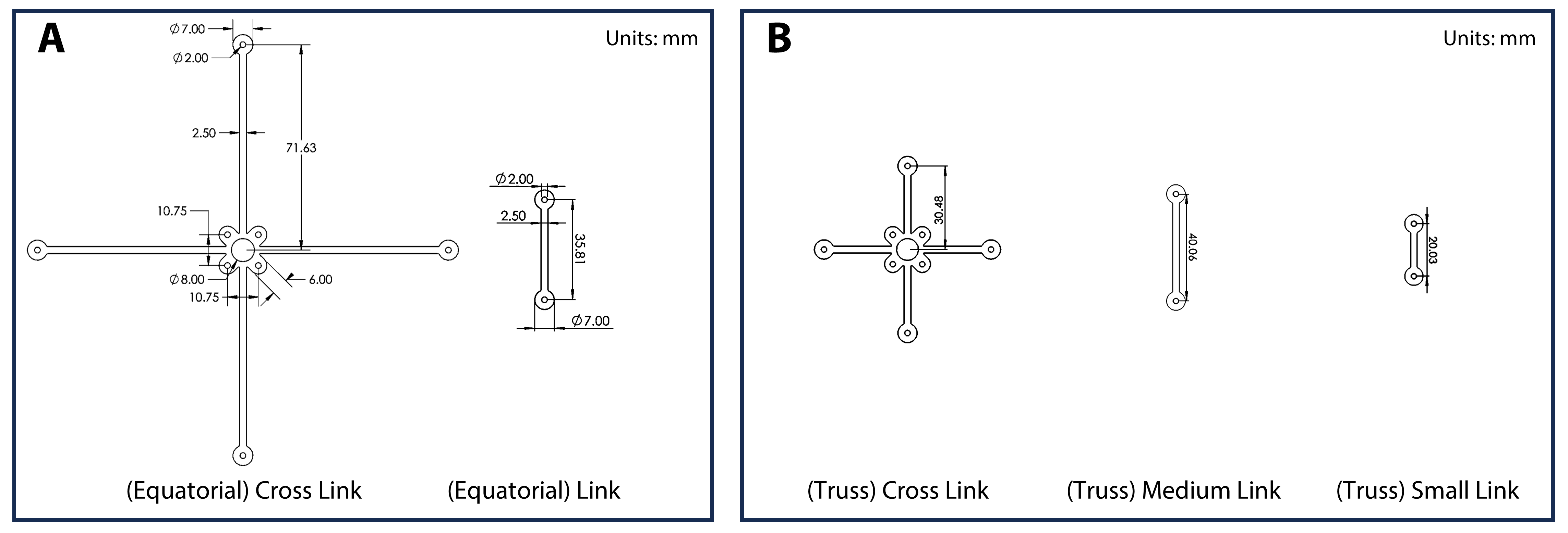}}
\caption{{Fig. S4. \bf Spring steel link dimensions.} Dimensions for the 1095 Spring Steel links used in the \textbf{(A)} equatorial and \textbf{(B)} truss TRUNCs.} 
\label{fig:links}
\end{figure}

Throughout our testing, we recorded the arm visiting a total of 137,300 poses randomly distributed over the workspace. After every 100 poses, the arm was cyclically compressed ($\Delta L = -70$ mm) five times. This corresponds to 6,865 compression cycles with a strain of $17.8\%$ and $11.8\%$ for the truss and equatorial cells, respectively. During the arm's training data collection, we only observed cells breaking when pushed substantially beyond bend angles of $\beta = 45 \degree$. In these extremal pose cases, we observed the ``cross link" (Fig. S\ref{fig:links}A) on the truss cell breaking near the corner where the link connects to the base. To mitigate these failures, we set the maximum cable retraction for the shoulder, elbow, and wrist to be -250 mm. In the future, the TRUNC's durability under large deflections could be improved by adding structural fillets to the cross-links to reduce stress concentration.

\subsection*{Configuration space sampling}

We sampled the arm's configuration space by randomly generating cable displacement values from the home position. Using an approximated model of the arm's mechanics, we generated separate cable displacement values for bending and compression. This model assumes that compression results from uniformly displacing all three cables at a segment whereas rotation results from non-uniformity among cable displacements. The displacement values were sampled using upper bounds on extension ($\Delta l_{\text{max}} = 60$) and rotation ($\Delta r_{\text{max,shoulder}} = 60$, $\Delta r_{\text{max,elbow}} = 60$, and $r_{\text{max,wrist}} = 80$). For sample $n$ configurations, we first generate an extension/contraction vector for the entire arm. 
\begin{equation*}
\Delta L = -\Delta l_{\text{max}} \cdot \text{rand}(n, 1)
\end{equation*}
To control rotation for the wrist, we generated random displacement values for all three cables and then randomly assigned one of the inputs to zero. 
\begin{align*}
\Delta R_{\text{wrist}} &= -\Delta r_{\text{max,wrist}} \cdot \text{rand}(n, 3) \\
\text{for } i &= 1:n, \quad \Delta R_{\text{wrist}}(i, \text{randi}(3)) = 0
\end{align*}
The rotation of the elbow joint follows a similar pattern:
\begin{align*}
\Delta R_{\text{elbow}} &= -\Delta r_{\text{max,elbow}} \cdot \text{rand}(n, 3) \\
\text{for } i &= 1:n, \quad \Delta R_{\text{elbow}}(i, \text{randi}(3)) = 0
\end{align*}
We coupled the shoulder and elbow rotations at this stage:
\begin{equation*}
\Delta R_{\text{shoulder}} = \Delta R_{\text{elbow}}
\end{equation*}
An offset is applied to $\Delta  L$ based on the amount of bending the arm experiences. Each joint compresses during rotation since the cables are in tension. To compensate for this, we slacken the cables proportionately to the max of $\Delta R$ for each joint:
\begin{equation*}
\Delta L_{\text{offset}} = \frac{\max(\lvert\Delta R_{\text{wrist}}\rvert) + \max(\lvert\Delta R_{\text{elbow}}\rvert) + \max(\lvert\Delta R_{\text{shoulder}}\rvert)}{\alpha}
\end{equation*}
\begin{equation*}
\Delta L = \Delta L + \Delta L_{\text{offset}}
\end{equation*}
We empirically determined that using a value of $\alpha=8$ could increase the arm's range of motion for large values of $\Delta L$

The segment displacements are then calculated by summing the $\Delta R$ and $\Delta L$ displacements. We scale $\Delta L$ based on the number of active and passive joints in each segment. 
\begin{align*}
\Delta \mathcal{S}_{\text{shoulder}} &= \Delta R_{\text{shoulder}} + \frac{3}{7} \cdot \Delta L \\
\Delta \mathcal{S}_{\text{elbow}} &= \Delta R_{\text{shoulder}} + \frac{2}{7} \cdot \Delta L \\
\Delta \mathcal{S}_{\text{wrist}} &= \Delta R_{\text{wrist}} + \frac{2}{7} \cdot \Delta L
\end{align*}
Finally, we calculate $\mathcal{L}$ by adding the displacements to the arm's home configuration. We empirically found that adding a compensation factor of $\beta = \frac{3}{4}$ to the wrist helped create more poses with the end effector normal to the ground. 
\begin{gather*}
    \Delta \mathcal{L} = \begin{bmatrix} \Delta \mathcal{S}_{\text{shoulder}}, \Delta \mathcal{S}_{\text{shoulder}} + \Delta \mathcal{S}_{\text{elbow}}, \beta \cdot (\Delta  \mathcal{S}_{\text{shoulder}} + \Delta \mathcal{S}_{\text{elbow}}) + \Delta  \mathcal{S}_{\text{wrist}} 
\end{bmatrix}^T \\
    \mathcal{L} = \mathcal{L}_{\text{home}} + \Delta \mathcal{L} 
\end{gather*}

\subsection*{Orientation subsets of workspace}

\begin{figure}[!h]
\makebox[\textwidth][c]{\includegraphics[width=\textwidth]{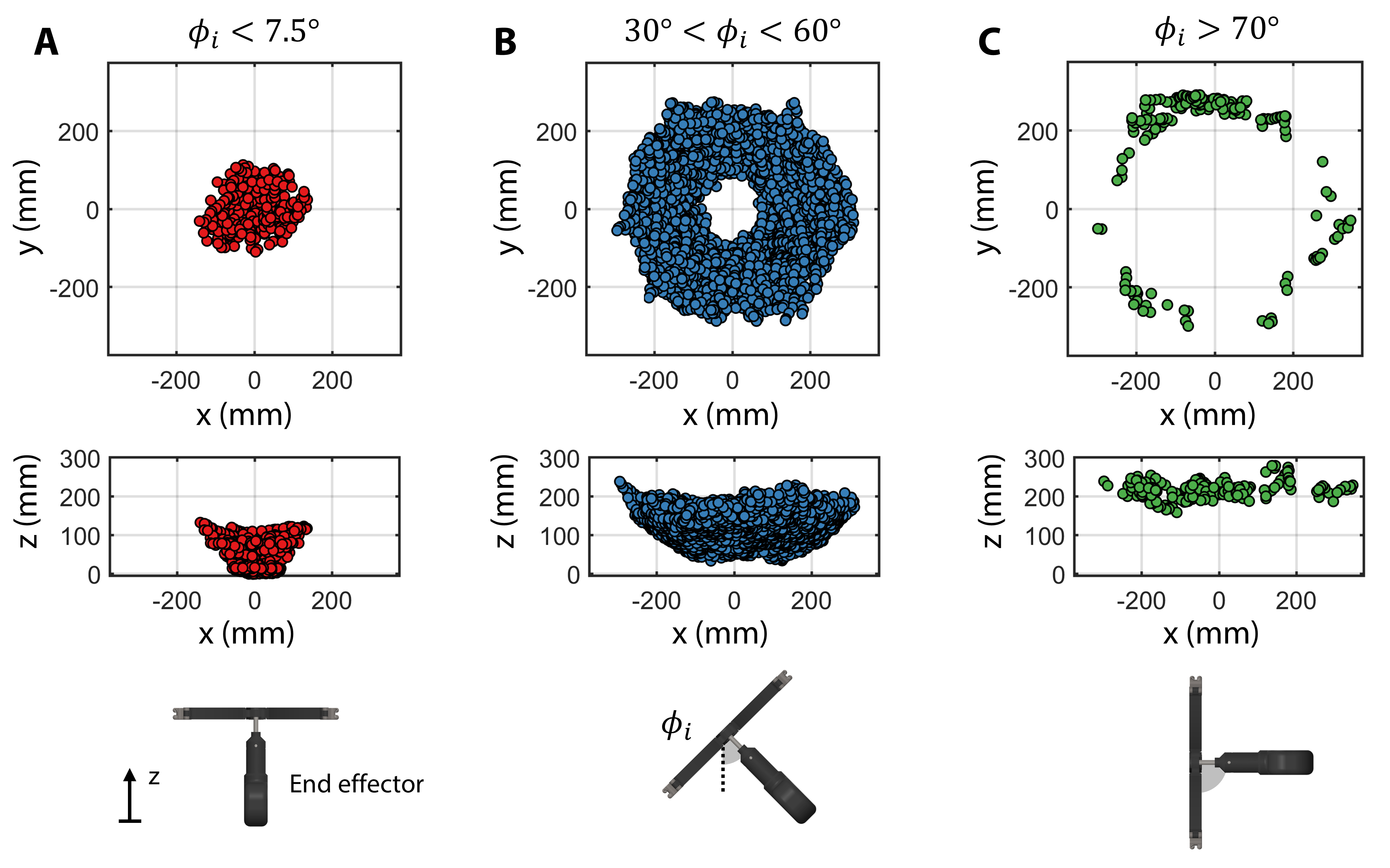}}
\caption{{Fig. S8. \bf Analysis of end effector orientations.} Regions of the workspace where the end effector is approximately {\bf (A)} perpendicular, {\bf(B)} partially tilted,  and {\bf(C)} parallel relative to the ground plane.} 
\label{fig:Sub-workspace}
\end{figure}

Robots frequently need to both position and orient their end effector when interacting with their environment. We analyzed the arm's workspace with constraints on orientation to find regions where the end effector was normal (A), partially tilted (B), and almost parallel to the ground (C) as shown in Fig. \ref{fig:Sub-workspace}. Given a set of quaternions $Q$, where each quaternion $q_i \in Q$ is represented as $q_i = (q_w, q_x, q_y, q_z)$, and the ground plane's normal vector is $\mathbf{u} = [0, 0, 1]^T$, we calculated the angle $\phi_i$ between the global z-axis and the axis defined by each quaternion's rotation of $\mathbf{u}$. First, we converted the quaternion $q_i$ to a 3x3 rotation matrix $R(q_i)$ using the quat2rotm function in MATLAB's Aerospace Toolbox. We then rotated the z-axis unit vector using the matrix derived from the quaternion
\[
\mathbf{v}_i = R(q_i) \cdot \mathbf{u}
\]
where $\mathbf{v}_i$ is the rotated vector. We then calculated the cosine of the angle between the rotated vector and the z-axis:
\[
\cos(\phi_i) = \frac{\mathbf{v}_i \cdot \mathbf{u}}{\|\mathbf{v}_i\| \cdot \|\mathbf{u}\|}
\]  
To prevent a domain error due to numerical inaccuracies, we clipped $\cos(\phi_i)$ to ensure its values were between $\begin{bmatrix}
    -1,1
\end{bmatrix}$:
\[
\cos(\phi_i) = \max(\min(\cos(\phi_i), 1), -1)
\]
Finally, we calculated the angular distance in radians:
    \[
    \phi_i = \arccos(\cos(\phi_i))
    \]
We calculated the angular distance for all quaternions in the set $Q$ and then filtered the workspace to find the regions of interest shown in Fig. S\ref{fig:Sub-workspace}. Within our workspace data, we found a clear correlation between $\phi_i$ and positional distance from the origin. Lifting the end effector to a parallel pose requires the arm to move towards the edge of the workspace while normal poses can be achieved closer to the origin.

\subsection*{Learning the inverse kinematics}

\begin{figure}
\makebox[\textwidth][c]{\includegraphics[width=\textwidth]{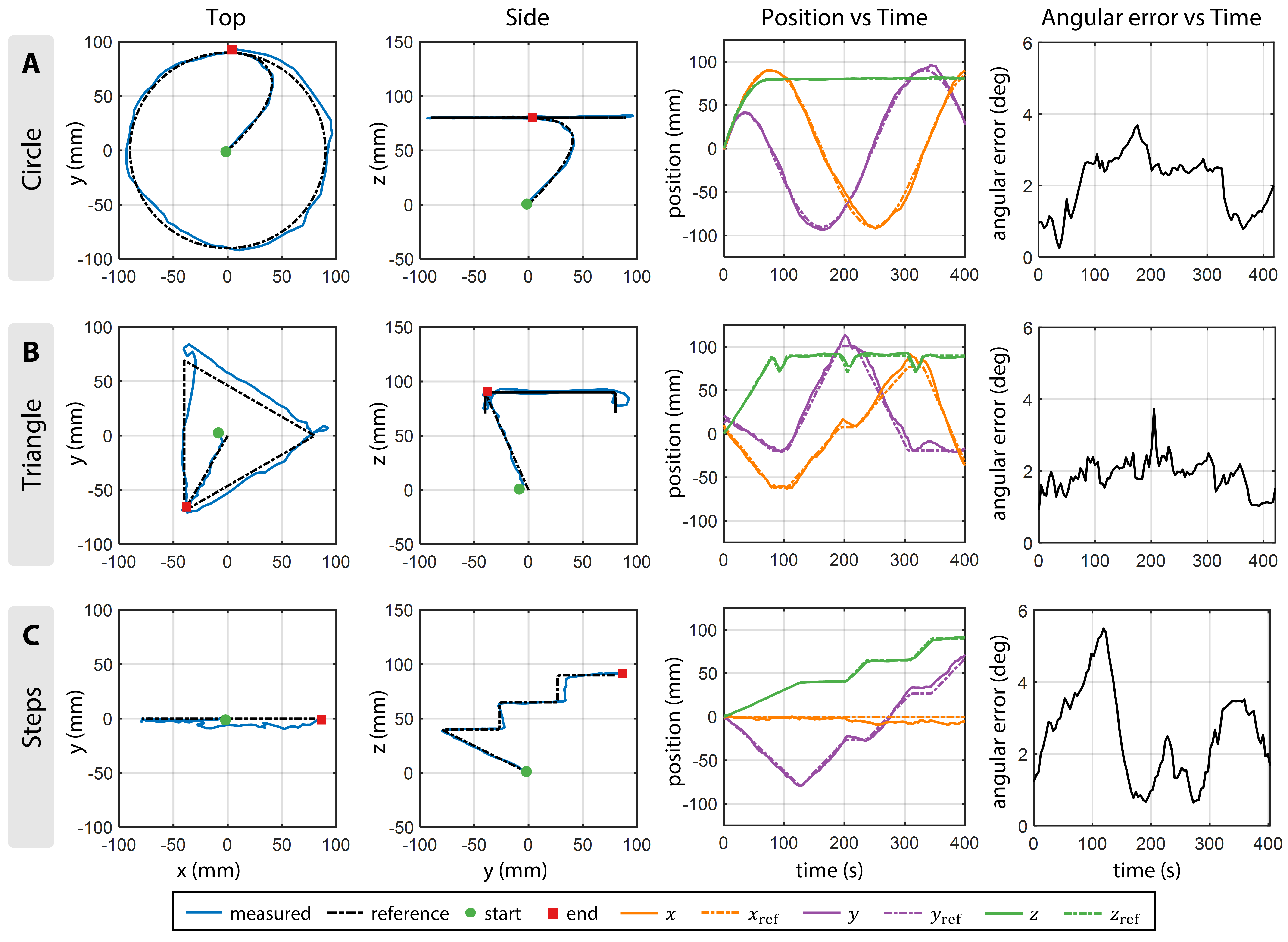}}
\caption{{Fig. \bf S9. Learning an inverse kinematic model for trajectory following.} We learned the arm's inverse kinematic mapping by training a DNN on a well-defined configuration space manifold. We then validated the model by following trajectories in the shape of {\bf (A)} a circle, {\bf (B)} triangle, and {\bf (C)} stairs. This is then compared to the ground truth motion over each trajectory to find the positional and angular error of the model.} 
\label{fig:Model}
\end{figure}

We trained a deep neural network (DNN) to learn the arm's inverse kinematics using the dataset of well-defined poses previously collected. These configurations represent a manifold of $\mathcal{L}$ where each cable is tensioned and the arm has full mobility. We evaluated the accuracy of our learned model using three representative trajectories designed to test performance across various movement patterns. For each test, the pose of the end effector was recorded at 100 reference waypoints with a $4$ second pause between movements (Fig. S\ref{fig:Model}). First, we tested a circular trajectory generated using cubic interpolation to create a smooth curve (Fig. S\ref{fig:Model}A). The arm tracked the reference input with a mean error of $5.0$ mm and $2.1 \degree$. Next, we generated a triangular trajectory using linear interpolation to create straight movements in the horizontal $xy$ plane and added retraction movements in the $-z$ direction at the triangle's vertices (Fig. S\ref{fig:Model}B). The mean error for this test was $7.3$ mm and $1.9 \degree$. The final test was a staircase path with linear movements in the $xz$ plane and abrupt changes in the $+z$ direction. The test had a mean error of $5.7$ mm and $2.5 \degree$ (Fig. S\ref{fig:Model}C). We note the arm is less sensitive to errors that map to the $z$ direction compared to the $x$ and $y$ directions. These results demonstrate our learned inverse kinematics model can accurately generate arbitrary trajectories within the arm's workspace, enabling basic motion planning for real-world tasks.

\newpage

\section*{Supplementary Figures}

\begin{figure}[!h]
\makebox[\textwidth][c]{\includegraphics[width=\textwidth]{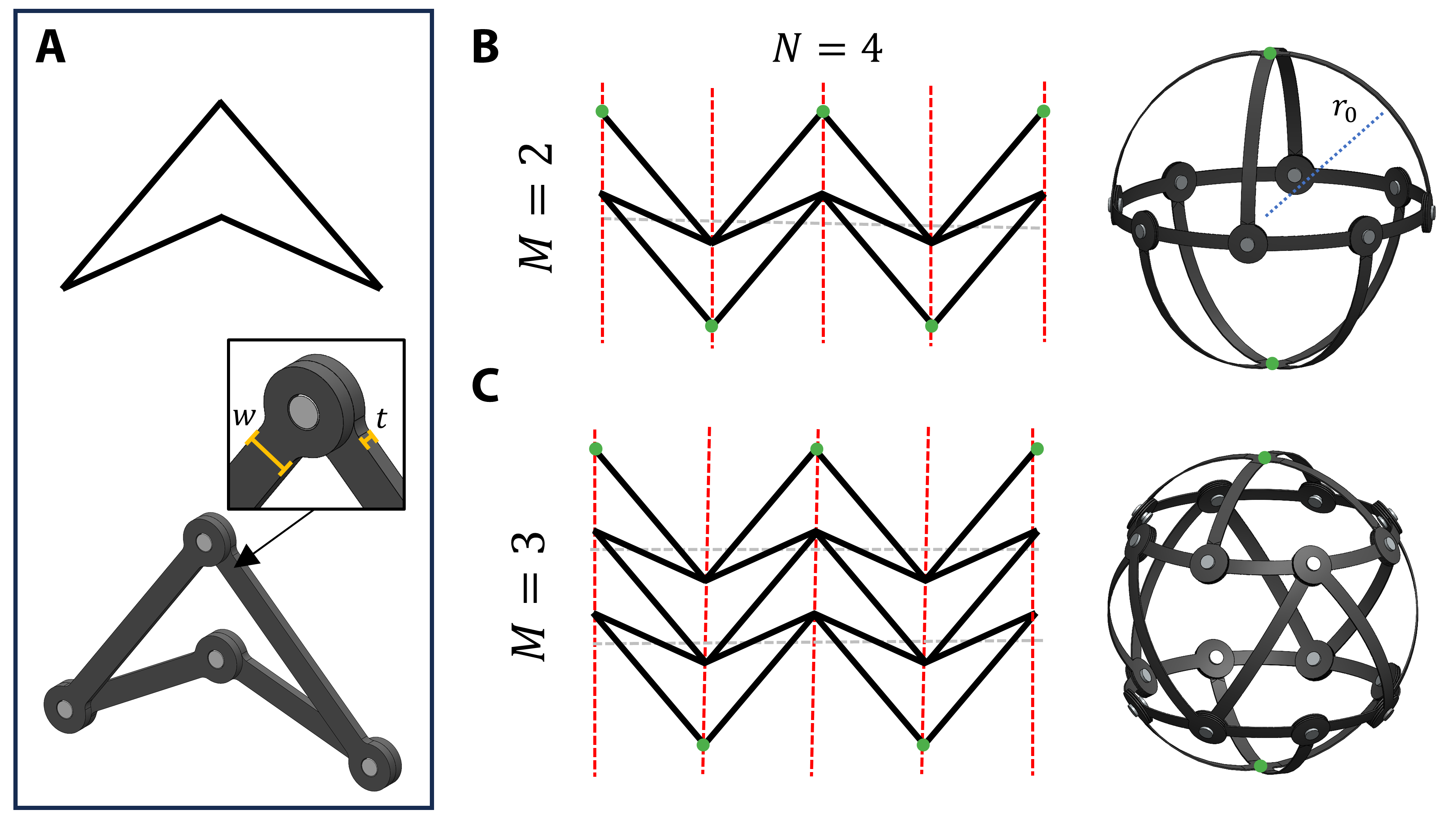}}
\caption{{Fig. S1. \bf TRUNCs are based on tilings of the double arrowhead.} {\bf (A) } The double arrowhead can be made as a linkage where links have a width $w$ and thickness $t$. {\bf (B)} The equatorial cell is based on a tiling where $N=4$ and $M=2$  while the truss cell ($N=4$ and $M=3$) has an additional row of arrow heads.}
\label{sup_fig:tiling}
\end{figure}

\clearpage

\subsection*{Supplementary Movies}

\textbf{Movie S1.} A TRUNC transmitting torque at various bend angles. \\
\textbf{Movie S2.} Nested and chained TRUNCs transmitting multiple independent torques. \\
\textbf{Movie S3.} The TRUNC arm installing a lightbulb.\\
\textbf{Movie S5.} The TRUNC arm fastening bolts alongside a human. \\
\textbf{Movie S5.} The TRUNC arm turning a valve. \\



\end{document}